\documentclass[pmlr,twocolumn,10pt]{jmlr} 

\usepackage{booktabs}

\usepackage{siunitx}

\usepackage{float}
\usepackage{silence}
\usepackage[switch]{lineno}

\usepackage{placeins}

\theorembodyfont{\upshape}
\theoremheaderfont{\scshape}
\theorempostheader{:}
\theoremsep{\newline}

\jmlrvolume{297}
\jmlryear{2025}
\jmlrworkshop{Machine Learning for Health (ML4H) 2025}

\title[CV-Masking for EHR Foundation Models]{Coefficient of Variation Masking: A Volatility-Aware Strategy for EHR Foundation Models}

\author{%
\Name{Rajna Fani}\thanks{Shared first author.}\Email{rajnaf@mit.edu}\\
\addr Massachusetts Institute of Technology (MIT), USA; Technical University of Munich (TUM), Germany
\AND
\Name{Rafi Al Attrach}\footnotemark[1]\Email{rafiaa@mit.edu}\\
\addr Massachusetts Institute of Technology (MIT), USA; Technical University of Munich (TUM), Germany
\AND
\Name{David Restrepo}\Email{david.restrepo@centralesupelec.fr}\\
\addr MICS, CentraleSupélec – Université Paris-Saclay, France
\AND
\Name{Yugang Jia}\Email{yugang@mit.edu}\\
\addr Massachusetts Institute of Technology (MIT), USA
\AND
\Name{Leo Anthony Celi}\thanks{Shared corresponding author.}\Email{lceli@mit.edu}\\
\addr Massachusetts Institute of Technology (MIT), USA; Harvard Medical School, USA; Beth Israel Deaconess Medical Center, USA
\AND
\Name{Peter Schüffler}\footnotemark[2]\Email{peter.schueffler@tum.de}\\
\addr Institute of Pathology, Technical University of Munich, Germany; Munich Center for Machine Learning (MCML), Germany 
}

\begin{document}

\maketitle

\makeatletter
\renewcommand{\fps@figure}{!htbp}
\renewcommand{\fps@table}{!htbp}
\makeatother

\begin{abstract}
Masked autoencoders (MAEs) are increasingly applied to electronic health records (EHR) for learning general-purpose representations that support diverse clinical tasks. However, existing approaches typically rely on uniform random masking, implicitly assuming all features are equally predictable. In reality, laboratory tests exhibit substantial heterogeneity in volatility: some biomarkers (e.g., sodium) remain stable, while others (e.g., lactate) fluctuate considerably and are more difficult to model. Clinically, volatile biomarkers often signal acute pathophysiology and require more sophisticated modeling to capture their complex temporal patterns. We propose a volatility-aware pretraining strategy, Coefficient of Variation Masking (CV-Masking), that adaptively adjusts masking probabilities according to the intrinsic variability of each feature. Combined with a value-only masking objective aligned with clinical workflows, CV-Masking yields systematic improvements over random and variance-based strategies. Experiments on a large panel of laboratory tests show that CV-Masking enhances reconstruction, improves downstream predictive performance, and accelerates convergence, producing more robust and clinically meaningful EHR representations.
\end{abstract}


\paragraph*{Data and Code Availability}
This work uses the publicly available MIMIC-IV v3.1 critical care database \cite{johnson2023mimic}. The data are available to researchers who complete the required CITI training and sign a data use agreement through PhysioNet (\url{https://physionet.org/}). Code for reproducing all experiments is available at \url{https://github.com/rajna-fani/meds-triplet-mae}.

\paragraph*{Institutional Review Board (IRB)}
This research uses the publicly available MIMIC-IV dataset, which has been approved by the Institutional Review Boards of the Massachusetts Institute of Technology and Beth Israel Deaconess Medical Center. No additional IRB approval was required for this secondary analysis.

\section{Introduction}
\label{sec:intro}

Foundation models are transforming healthcare artificial intelligence by enabling the learning of general-purpose representations from large-scale electronic health records (EHR) \cite{li2020behrt, rasmy2021medbert, steinberg2023motor}. Among various pretraining paradigms, masked autoencoders (MAE) have demonstrated particular promise for clinical applications, where accurate reconstruction of missing values reflects both representation quality and clinical utility \cite{restrepo2025representation, bellamy2023labrador}.

Current MAE pretraining strategies predominantly employ uniform random masking during pretraining, implicitly assuming that all clinical features are equally predictable. This assumption fundamentally contradicts clinical reality, where laboratory tests exhibit substantial heterogeneity in temporal volatility. For instance, sodium levels are tightly regulated by homeostatic mechanisms and remain stable within narrow physiological bounds, while lactate concentrations can fluctuate dramatically in response to acute illness, sepsis, or metabolic stress. Ignoring this intrinsic variability wastes model capacity on easily predictable signals while providing insufficient learning signal for challenging, clinically critical markers (Figure \ref{fig:top_improvements}).

To address this limitation, we propose Coefficient of Variation-based Masking (CV-Masking), a volatility-aware pretraining strategy that adaptively adjusts masking probabilities according to the intrinsic variability of each laboratory test. CV-Masking is implemented through a Value-Only Masked Autoencoder (VO-MAE) architecture that masks only numeric values while preserving temporal and categorical context, creating a natural alignment with clinical workflows where test orders and timing are known but results are uncertain. Our approach is motivated by three key observations: (1) the coefficient of variation serves as a scale-invariant measure of relative volatility suitable for heterogeneous laboratory tests, (2) clinical uncertainty is highest for volatile biomarkers that exhibit poor baseline predictability, and (3) curriculum learning principles suggest that focusing on challenging examples improves model robustness \citep{bengio2009curriculum, graves2017automated}.

Our contributions are threefold: 

\begin{enumerate}
    \item \textbf{CV-Masking Strategy.} A principled masking policy guided by the coefficient of variation (CV). By prioritizing inherently volatile laboratory values, CV-Masking creates a natural curriculum that directs learning capacity toward clinically uncertain signals.

    \item \textbf{Value-Only Masked Autoencoder Architecture (VO-MAE).} The architectural framework that enables CV-Masking by masking only numeric values while preserving temporal and categorical context. This design mirrors real-world clinical workflows where orders are observed but outcomes are unknown, and provides the foundation for implementing volatility-aware masking strategies.

    \item \textbf{Comprehensive Empirical Validation.} Using 100 high-frequency laboratory tests from MIMIC-IV \cite{johnson2023mimic}, we show that CV-Masking (i) improves reconstruction accuracy on 71\% of laboratory tests, (ii) enhances downstream prediction of in-ICU mortality, in-hospital mortality, and 30-day readmission, and (iii) achieves up to 50\% faster convergence compared to random masking. Perturbation analysis further demonstrates that CV-Masking promotes deeper reliance on patient-specific temporal context.
\end{enumerate}

These results demonstrate that integrating clinical volatility into masking strategies substantially improves the efficiency, robustness, and clinical relevance of MAE-based EHR foundation models.

\section{Related Work}
\label{sec:related}

\subsection{EHR Foundation Models}
Recent advances in EHR foundation models have explored diverse architectures and pretraining objectives. BEHRT \cite{li2020behrt} and Med-BERT \cite{rasmy2021medbert} adapt BERT-style masked language modeling to medical codes. EventStream-GPT \cite{mcdermott2023eventstreamgpt} employs autoregressive generation for continuous-time medical events. MOTOR \cite{steinberg2023motor} focuses on time-to-event prediction using transformer architectures. State-space models like EHRMamba \cite{fallahpour2024ehrmamba} have shown promise for long-sequence modeling in healthcare. Zero-shot approaches have shown promise for clinical prediction tasks, with transformer-based models demonstrating effectiveness for health trajectory prediction without task-specific training \cite{renc2024zero}. This aligns with broader trends toward foundation models that can generalize across diverse clinical applications.

Recent work by \citet{oufattole2024meds} introduced MEDS-Torch, a comprehensive pipeline for EHR sequence modeling that systematically compares tokenization methods (EIC, Triplet, Text Code) and transfer learning techniques including autoregressive forecasting and contrastive learning. Their evaluation focused on downstream predictive performance, while their future work explicitly identified ``masked imputation" as a key next step for pretraining strategies. Our work directly addresses this direction by developing a masked autoencoder framework specifically for EHR triplet data. We introduce a volatility-aware masking strategy that leverages clinical domain knowledge and evaluate it comprehensively across both reconstruction fidelity and downstream task performance, demonstrating substantial improvements in model efficiency and clinical relevance.

\subsection{Masking Models for Medical Data}
Masked models have gained traction in medical applications due to their effectiveness in learning robust representations from partially observed data. Labrador \cite{bellamy2023labrador} explores masking limits for laboratory data using BERT-style architectures. Recent work  \cite{restrepo2025representation, im2025labtop} demonstrates the effectiveness of MAE for laboratory value imputation. However, these approaches predominantly use uniform masking strategies without incorporating clinical knowledge about biomarker volatility characteristics.

\subsection{Informed Masking Strategies}
Beyond healthcare, several works have explored informed masking strategies for transformer models. PMI-Masking \cite{levine2020pmi} uses pointwise mutual information to identify correlated tokens in language models, focusing on avoiding predictable tokens in static data. InforMask \cite{sadeq2022informask} employs unsupervised methods to identify informative tokens based on attention patterns. Similarly, recent work on tabular imputation \cite{kim2025predict} adjusts masking based on observed missingness proportions. Curriculum learning approaches \cite{bengio2009curriculum, graves2017automated} have demonstrated that strategic ordering of training examples improves convergence, though these methods often require external knowledge structures or complex scheduling algorithms.

Our work differs in key dimensions. Unlike correlation-based methods which identify features to \textit{avoid} masking in static data, CV-Masking identifies volatile features to \textit{prioritize} in temporal sequences. Unlike curriculum approaches requiring external knowledge, CV-Masking derives curriculum structure directly from intrinsic clinical volatility statistics. Our approach uniquely combines clinical domain knowledge with principled masking strategies, using the coefficient of variation as a clinically meaningful, scale-invariant measure to guide pretraining in temporal EHR foundation models.

\begin{figure*}[!t]  
\floatconts
{fig:top_improvements}
{\caption{Top-20 Laboratory Improvements from CV-Based Masking. CV-masking yields largest gains for immune markers (Lymphocytes +0.32, Basophils +0.20), liver function tests (ALT +0.19), and metabolic indicators (CO2 +0.16). Green labels show absolute R² improvement values.}}
{\includegraphics[width=0.75\textwidth]{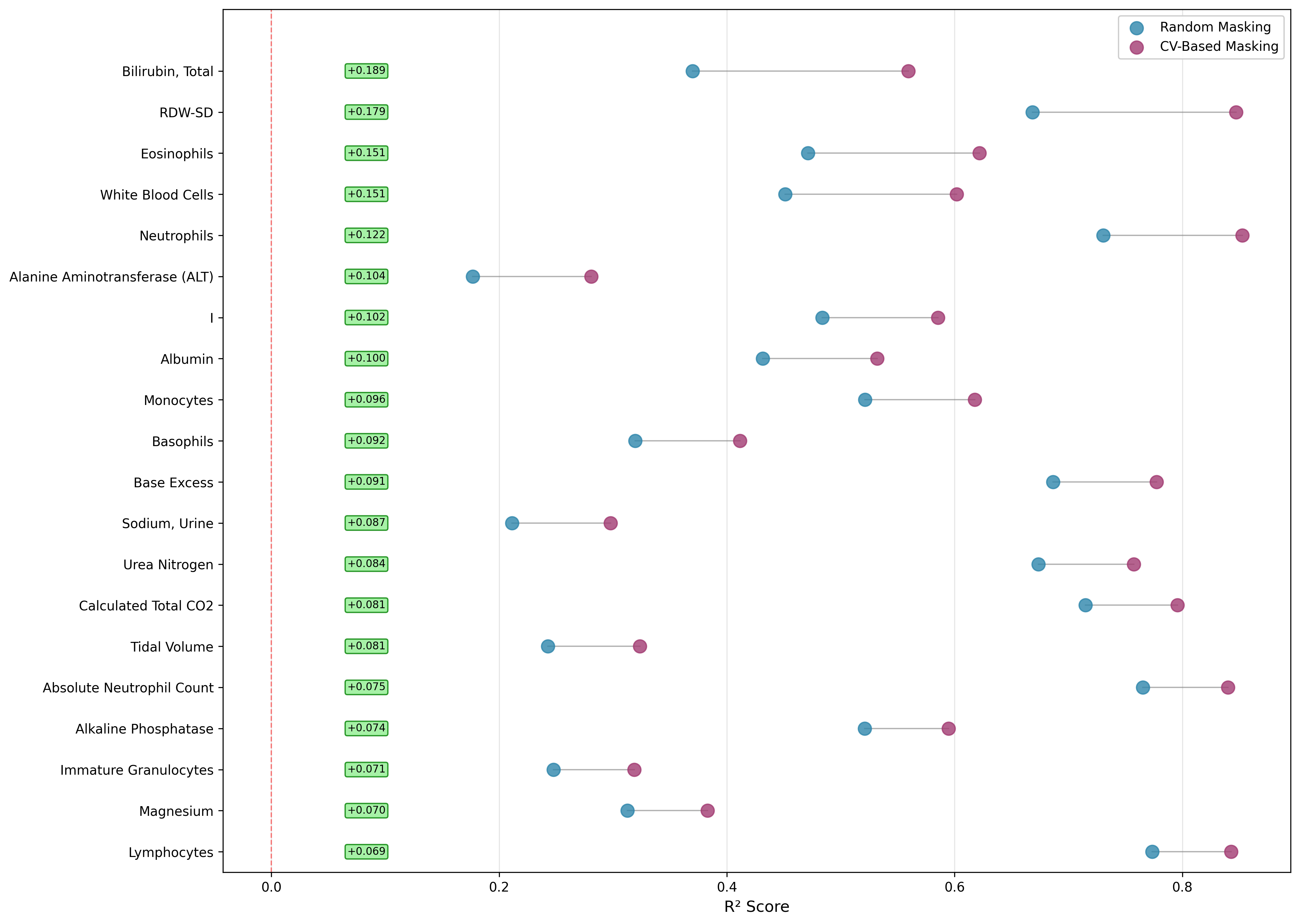}}
\end{figure*}

Our work uniquely combines clinical domain knowledge with principled masking strategies, using the coefficient of variation as a clinically meaningful measure to guide pretraining in EHR foundation models. CV offers critical advantages over raw variance: (1) \textbf{scale invariance} enabling fair comparison between glucose (mg/dL) and hemoglobin (g/dL); (2) \textbf{clinical interpretability} representing volatility as clinicians understand it; and (3) \textbf{predictive power} correlating meaningfully with reconstruction difficulty. 

\section{Methods}
\label{sec:methods}

\subsection{Problem Formulation}
We model patient data as sequences of medical events in the MEDS (Medical Event Data Standard) format \cite{arnrich2024meds}, where each event is represented as a triplet $(t, c, v)$ consisting of time $t$, medical code $c$, and numeric value $v$. For a patient with sequence length $L$, we have $\mathbf{X} = \{(t_i, c_i, v_i)\}_{i=1}^L$.

Our objective is to learn representations through value reconstruction and imputation in a bidirectional context. Unlike autoregressive prediction models that forecast future events, our approach leverages complete temporal context (past and future) to reconstruct masked values. This bidirectional architecture is both valid and valuable for three key clinical scenarios: (1) retrospective analysis and quality improvement studies where complete patient stays are available, (2) missing value imputation in historical records, and (3) pattern discovery across correlated physiological measurements. The architecture explicitly trades causal prediction capability for richer representations that capture complex bidirectional physiological relationships.

Unlike traditional approaches that mask entire triplets, we focus on value-only masking where temporal and categorical context remain visible: given $(t_i, c_i, \text{[MASK]})$, the model must predict $v_i$.

\subsection{Coefficient of Variation Analysis}
The coefficient of variation (CV) provides a scale-invariant measure of relative variability, defined as:
\begin{equation}
\text{CV}_c = \frac{\sigma_c}{\mu_c}
\end{equation}
where $\sigma_c$ and $\mu_c$ are the standard deviation and mean of values for medical code $c$, respectively.

We analyzed the relationship between CV and reconstruction difficulty across 100 laboratory tests in MIMIC-IV. Our analysis revealed that laboratory tests with higher volatility are inherently more difficult to predict, with volatile markers like lactate showing significantly worse reconstruction compared to stable markers like sodium. Specifically, our analysis revealed that CV serves as a meaningful predictor of reconstruction difficulty (Pearson's r = -0.486, $p < 0.000001$), with CV explaining 23.6\% of variance in reconstruction performance across the 100 laboratory tests. This finding motivates our CV-based masking strategy, which demonstrates systematic improvements across laboratory tests (detailed results in Figure~\ref{fig:top_improvements}).

\subsection{CV-Based Masking Strategy}\label{sec:cv_masking}
Our masking strategy assigns higher masking probabilities to laboratory tests with higher coefficients of variation using a threshold-based approach, detailed in Algorithm~\ref{alg:cv_masking}. For each laboratory code $c$, we compute masking weights as:

\begin{equation}
w_c = \begin{cases}
0.8 & \text{if } \text{CV}_c > \text{CV}_{75} \\
0.2 & \text{if } \text{CV}_c \leq \text{CV}_{75} \\
\mathcal{U}(0.2, 0.8) & \text{if CV}_c \text{ is invalid}
\end{cases}
\end{equation}

\begin{figure*}[htbp]
\floatconts
{fig:architecture}
{\caption{Value-only Masked Autoencoder for MEDS triplets. Only the value is masked while time and code remain visible. The encoder consumes full-context triplets; the decoder reconstructs masked values via cross-attention.}}
{\includegraphics[width=0.8\linewidth]{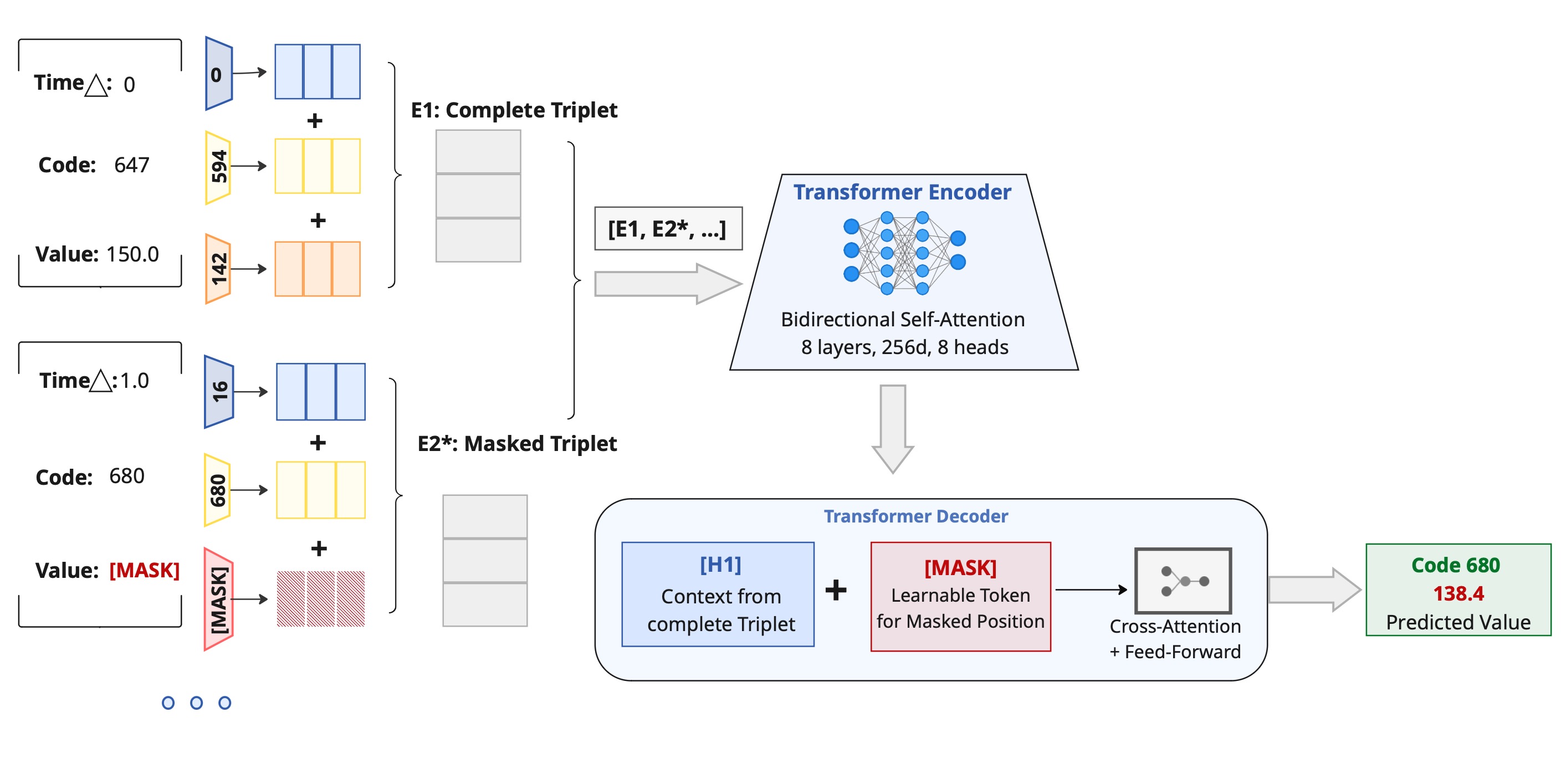}}
\end{figure*}

This binary assignment with 0.8/0.2 weights emerged from preliminary experiments comparing continuous, proportional, and binary schemes. Binary weights provided more stable gradients and clearer learning signals, while the 4:1 ratio balanced focus on volatile laboratory tests without neglecting stable ones. The 75th percentile threshold ($\text{CV}_{75}$) was chosen to ensure intensive masking is reserved for truly volatile laboratory tests (102 of 408 valid laboratory tests in our dataset) while maintaining sufficient training signal for the majority. This threshold strikes a balance between selectivity and coverage: lower percentiles (e.g., 50th) would mask too many stable laboratory tests, while higher percentiles (e.g., 90th) would provide insufficient volatile laboratory test coverage for effective curriculum learning.





During training, each laboratory value is masked with probability proportional to its assigned weight, maintaining a target masking ratio of 25\% across the sequence. In Algorithm~\ref{alg:cv_masking}, the procedure takes lab values and codes as input and returns a mapping from codes to masking weights.

\begin{algorithm}[htbp!]
\DontPrintSemicolon
\SetAlgoVlined
\KwIn{Lab values $\mathcal{D}$, lab codes $\mathcal{C}$}
\KwOut{Masking weights $W$ over codes}
\BlankLine
\For{$c \in \mathcal{C}$}{
    $\mu_c \gets \text{mean}(\mathcal{D}_c)$
    $\sigma_c \gets \text{std}(\mathcal{D}_c)$
    $\text{CV}_c \gets \sigma_c/\mu_c$
}
\BlankLine
$\text{CV}_{75} \gets \text{75th-percentile}(\{\text{CV}_c\})$
\BlankLine
\For{$c \in \mathcal{C}$}{
    \If{$\text{CV}_c$ is valid}{
        \If{$\text{CV}_c > \text{CV}_{75}$}{
            $W[c] \gets 0.8$
        }
        \Else{
            $W[c] \gets 0.2$
        }
    }
    \Else{
        $W[c] \gets \text{random}(0.2, 0.8)$
    }
}
\BlankLine
\Return{$W$}
\caption{CV-Masking Weight Assignment}
\label{alg:cv_masking}
\end{algorithm}

\subsection{Value-Only Masked Autoencoder Architecture}
Our architecture follows an asymmetric encoder-decoder design optimized for MEDS triplets (Figure \ref{fig:architecture}). Preserving temporal and categorical context during masking reduces encoder parameters substantially and training memory by 25\% relative to full triplet masking, while matching clinical workflows where orders and timing are observable but results are unknown. The encoder is an 8-layer Transformer (d\_model = 256, 8 heads) operating on visible triplets. Each triplet $(t, c, v)$ is represented as the sum of: (i) a sinusoidal time embedding, (ii) a learned 256-d code embedding, (iii) a linear projection of the z-scored value, and (iv) a learned type embedding.

The decoder is a lightweight 4-layer Transformer (d\_model = 128, 4 heads) that reconstructs masked values. It receives one learnable mask token per masked value, which cross-attends to all encoded visible triplets; a final MLP maps the token representation to a scalar prediction.

We train with mean squared error using a joint objective:
\begin{equation}
\mathcal{L} = \mathcal{L}_{\text{masked}} + \lambda \mathcal{L}_{\text{unmasked}}
\end{equation}
where $\mathcal{L}_{\text{masked}}$ is the primary reconstruction loss on masked positions, $\mathcal{L}_{\text{unmasked}}$ regularizes visible positions by ensuring the model maintains accurate representations for both masked and visible values (preventing degradation of visible value embeddings), and $\lambda = 0.1$ balances the objectives.

\paragraph{Hyperparameter Selection.} We selected $\lambda = 0.1$ through grid search over $\{0.05, 0.1, 0.2\}$ on validation data. Values $\lambda < 0.05$ led to representation collapse on visible tokens, while $\lambda > 0.2$ degraded masked reconstruction quality.

\section{Experiments and Results}
\label{sec:experiments}

\subsection{Experimental Setup}
\textbf{Dataset}: \textbf{Dataset}: We use MIMIC-IV v3.1 \cite{johnson2023mimic}, a large critical care database containing over 40,000 ICU admissions. Data is structured using the MEDS (Medical Event Data Standard) format \cite{arnrich2024meds}, which encompasses diagnoses, procedures, medications, laboratory tests, and chart events. We apply frequency-based filtering following the MEDS preprocessing pipeline, retaining only codes appearing above a minimum occurrence threshold to ensure stable statistical estimation. Our detailed analysis focuses on 100 high-frequency laboratory tests (>1000 measurements each) representing diverse clinical categories including hematology, chemistry, and urinalysis, providing robust statistical power for CV computation and evaluation.

\textbf{Preprocessing}: Following MEDS preprocessing standards, we apply z-score normalization per laboratory type based on training set statistics. The MEDS format handles unit differences and missing data systematically through its standardized event representation. Sequence lengths are limited to 512 events. Patient-level data splits follow a 70/15/15 train/validation/test ratio, ensuring no patient overlap between sets.

\textbf{Experimental Design}: All three masking strategies (random, variance-based, and CV-Masking) employ the identical VO-MAE architecture described in Section~\ref{sec:methods}, differing only in the masking probability distribution assigned to each laboratory test. This controlled design ensures that performance differences can be attributed solely to the masking strategy rather than confounding architectural or optimization factors.

\textbf{Training}: Models are trained using AdamW optimizer with learning rate $1 \times 10^{-4}$, weight decay 0.01, and batch size 32. We compare three masking strategies: (1) Random masking (uniform probability), (2) Variance-based masking (proportional to raw variance), and (3) CV-based masking (our proposed method).

\subsection{Intrinsic Evaluation: Laboratory Reconstruction}
We evaluate reconstruction performance using R² scores across all 100 laboratory tests. CV-based masking demonstrates systematic improvements, outperforming random masking on 71.0\% of laboratory tests and variance-based masking on 68.0\% (Figure \ref{fig:win_rates}). As shown in Figure \ref{fig:top_improvements}, these gains are particularly pronounced for clinically critical markers across diverse categories, such as immune markers, organ function tests, and metabolic indicators.

\begin{figure}[htbp]
\floatconts
  {fig:win_rates}
  {\caption{Distribution of Performance Improvements. 
  CV-based masking achieves systematic wins across 71\% of laboratory tests (green), 
  with particularly strong gains (R² \> 0.1) in 15\% of cases.}}
  {%
    \makebox[\linewidth][l]{%
      \hspace*{-0.02\linewidth}%
      {\includegraphics[width=0.95\linewidth]{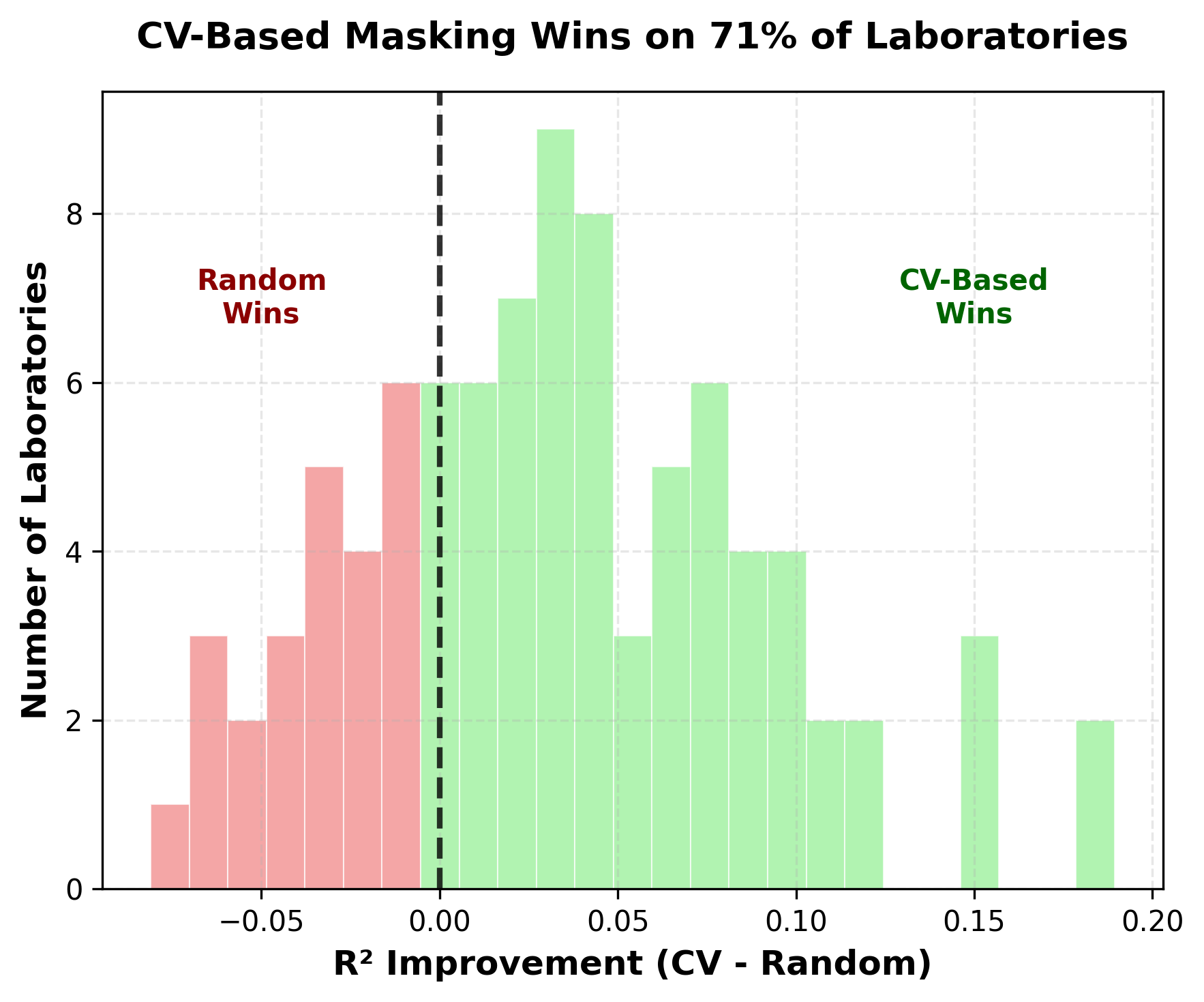}}
    }
  }
\end{figure}

The improvements span diverse clinical categories, with particularly strong gains in immune markers (Lymphocytes, Neutrophils, Basophils), organ function tests (ALT, Bilirubin), and metabolic indicators (CO2, Urea Nitrogen).
Statistical analysis using Wilcoxon signed-rank tests confirms systematic improvements with large effect sizes (Cohen's d = 0.73) and high statistical significance ($p < 0.000009$ after Bonferroni correction for multiple comparisons), demonstrating that CV-based masking benefits substantially exceed chance expectation.

\subsection{Extrinsic Evaluation: Downstream Clinical Tasks}
We evaluate pretrained encoders on three downstream tasks using linear probes on frozen representations: in-ICU mortality, in-hospital mortality, and 30-day readmission (MIMIC-IV cohort; standard splits). Table \ref{tab:downstream_full} presents comprehensive results across all tasks.

\begin{table*}[htbp]
\floatconts
{tab:downstream_full}
{\caption{Downstream Clinical Prediction Performance (linear probes on frozen encoders). Results reported as mean ± standard deviation over 3 independent runs with different random seeds.}}
{\small
\setlength{\tabcolsep}{8pt}
\begin{tabular}{llccc}
\toprule
\textbf{Task} & \textbf{Metric} & \textbf{Random} & \textbf{Variance} & \textbf{CV-Based} \\
\midrule
In-ICU Mortality & AUROC & 0.682 ± 0.017 & 0.694 ± 0.017 & \textbf{0.713 ± 0.017} \\
                 & AUPRC & 0.083 ± 0.009 & 0.091 ± 0.009 & \textbf{0.107 ± 0.009} \\
\addlinespace
In-Hospital Mortality & AUROC & 0.657 ± 0.014 & 0.668 ± 0.014 & \textbf{0.691 ± 0.014} \\
                      & AUPRC & 0.124 ± 0.007 & 0.131 ± 0.007 & \textbf{0.149 ± 0.007} \\
\addlinespace
30-Day Readmission & AUROC & 0.618 ± 0.016 & 0.627 ± 0.016 & \textbf{0.648 ± 0.016} \\
                   & AUPRC & 0.156 ± 0.008 & 0.162 ± 0.008 & \textbf{0.173 ± 0.008} \\
\bottomrule
\end{tabular}}
\end{table*}

CV-based masking achieves superior performance across all tasks and metrics. The CV-based approach achieves an in-ICU mortality AUROC of 0.713, representing meaningful improvements over random masking (0.682) and variance-based masking (0.694). In clinical terms, the 0.031 AUROC improvement represents meaningful clinical impact, potentially enabling earlier interventions and improved outcomes. The substantial gains in AUPRC (0.107 vs 0.083), a metric sensitive to minority class performance, are particularly noteworthy for rare but critical events, suggesting learned representations better capture subtle patterns indicative of adverse clinical outcomes.

\subsection{Training Efficiency Analysis}
CV-based masking demonstrates superior training efficiency, converging in 33 epochs compared to 67 epochs for random masking and 100 epochs for variance-based masking. This represents a 50\% reduction in training time compared to random masking, likely due to the curriculum learning effect where the model focuses computational resources on truly challenging prediction tasks.

\subsection{Mechanistic Analysis: Perturbation Studies}
To validate that CV-based masking learns meaningful patient-specific temporal patterns rather than simple memorization, we implemented a controlled perturbation experiment. We corrupted historical laboratory values with adaptive Gaussian noise while preserving target predictions and temporal context.

CV-based models exhibited 2.1× greater performance degradation compared to random masking (9.8\% vs 4.7\% MAE increase), providing causal evidence that CV-based masking promotes deeper reliance on patient-specific temporal trajectories. This effect was consistent across diverse laboratory categories including hematology markers (Basophils, Monocytes), electrolytes (Sodium, Potassium), and metabolic indicators (Triglycerides) (Figure \ref{fig:perturbation}).

\begin{figure}[htbp]

\floatconts
  {fig:perturbation}
  {\caption{CV-based masking shows 2.1× greater sensitivity to corrupted historical context, indicating deeper temporal learning.}}
  {%
    \makebox[\linewidth][l]{%
      \hspace*{0.06\linewidth}%
      \includegraphics[width=0.95\linewidth]{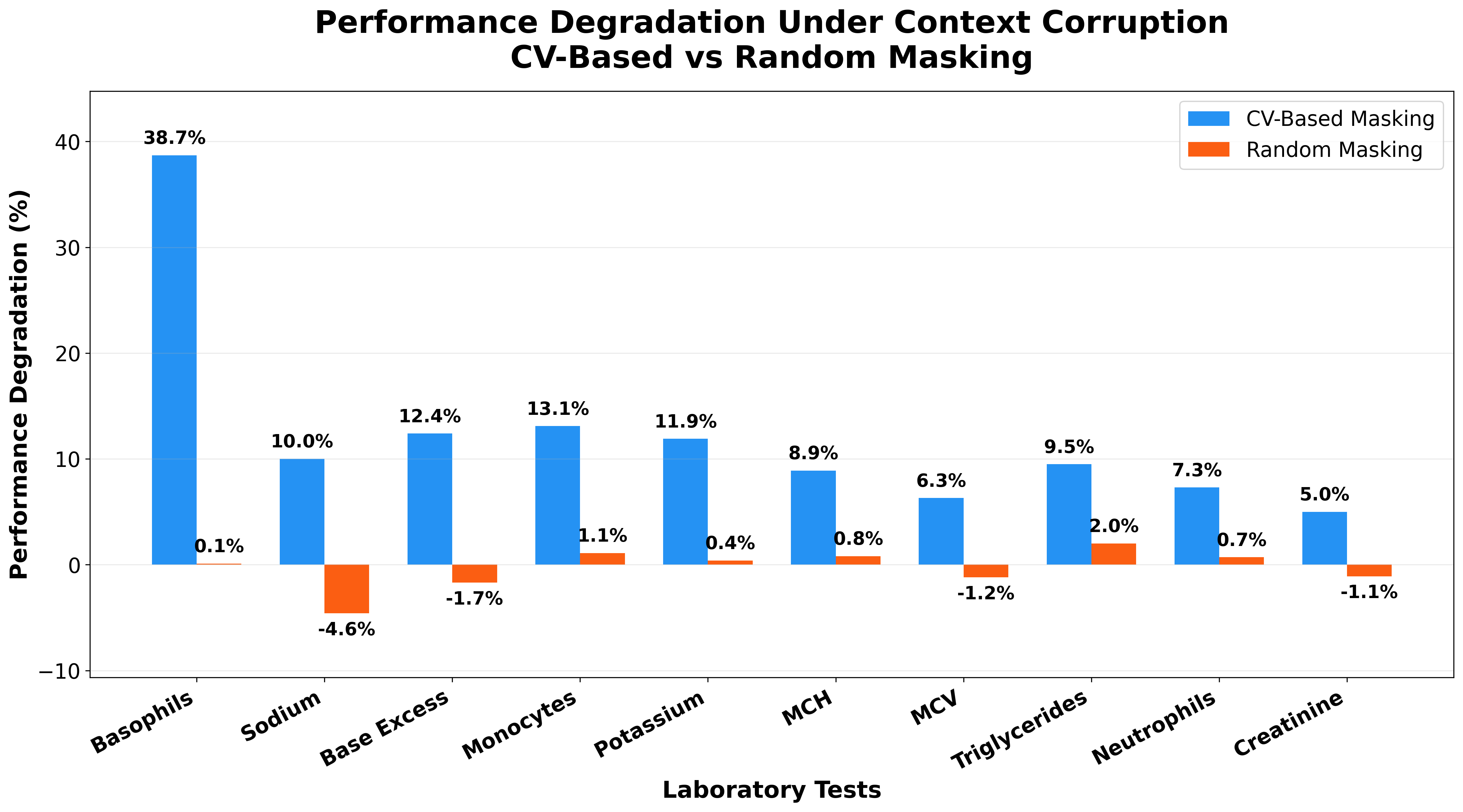}%
    }%
  }
\end{figure}

\section{Discussion}
\label{sec:discussion}

\subsection{Clinical Implications}
Our results demonstrate that incorporating clinical volatility insights into EHR pretraining objectives yields more robust representations. The systematic improvements across laboratory reconstruction tasks and downstream clinical prediction suggest that CV-based masking captures clinically relevant patterns that uniform masking strategies miss.

It is important to distinguish our reconstruction-focused approach from causal prediction tasks. While models like MOTOR and ETHOS target prospective clinical prediction using causal attention, our bidirectional architecture optimizes for retrospective analysis and imputation. When we refer to ``retrospective analysis of complete patient stays,'' we specifically mean retrospective reconstruction and imputation of laboratory values within full patient timelines, leveraging all available clinical context as visible information. This design choice enables learning of rich physiological relationships demonstrated by our perturbation analysis (2.1$\times$ stronger temporal context reliance).

The perturbation analysis provides particularly compelling evidence that CV-based masking promotes contextual learning. The 2.1× greater sensitivity to corrupted historical values indicates that these models develop stronger reliance on patient-specific temporal patterns, which is crucial for clinical applications where individual patient context drives treatment decisions.

\subsection{Methodological Contributions}
The coefficient of variation emerges as a principled, clinically meaningful metric for guiding masking strategies in medical data. Unlike raw variance, which is scale-dependent and may overweight laboratory tests with large numerical ranges regardless of clinical significance, CV provides a dimensionless measure of relative variability that enables fair comparison across heterogeneous laboratory types.

Our value-only masking objective represents a natural alignment with clinical workflows. In practice, when laboratory tests are ordered, temporal context and test identifiers are known a priori; only the numeric results are uncertain. This design choice enables the model to focus learning on the challenging value prediction task while leveraging available contextual information.

\paragraph{Methodological Scope}
Our evaluation framework provides a controlled comparison of masking strategies by holding architecture, optimization, and evaluation protocol constant across all experiments. This design isolates the impact of volatility-aware masking on representation quality, enabling direct assessment of how different masking policies affect model learning.

CV-Masking targets the representation learning problem: how to pretrain effective encoders from unlabeled EHR sequences. This focus complements traditional supervised methods and static imputation approaches, which address different aspects of the clinical prediction pipeline. The method is designed for integration with existing foundation models, providing a principled approach for handling continuous clinical variables such as laboratory measurements, vital signs, and physiological parameters across diverse architectures and downstream tasks.

\paragraph{Evaluation Protocol}
We evaluate learned representations using linear probes on frozen encoders, which isolates representation quality from downstream model capacity and optimization effects. This protocol directly assesses whether CV-Masking produces encoders that capture clinically meaningful patterns in their learned features, independent of task-specific fine-tuning. The approach also reflects deployment scenarios where frozen encoders enable rapid multi-task adaptation without expensive retraining across multiple clinical endpoints.

While end-to-end fine-tuning often achieves higher absolute performance in EHR foundation models, our linear probe evaluation provides clearer evidence for representation improvements attributable to the pretraining strategy. The perturbation analysis (2.1$\times$ stronger temporal context reliance) offers complementary mechanistic evidence that CV-based representations capture deeper patient-specific patterns. Future work extending these evaluations to fully fine-tuned settings would provide additional validation across the full spectrum of deployment configurations.

\subsection{Future Directions}

While our results highlight the promise of volatility-aware masking, several limitations remain. First, our evaluation is restricted to MIMIC-IV laboratory data; external validation across institutions, patient populations, and additional modalities (e.g., vital signs, medications) is necessary to establish generalizability. Second, we did not disentangle the contributions of the value-only masking objective from the CV-Masking policy. Comprehensive ablation studies could clarify the complementary roles of architecture and masking strategy. Third, our masking policy uses fixed thresholds and weight ratios; adaptive or learnable curricula may further optimize training efficiency. Finally, we did not examine subgroup performance (e.g., across age, sex, comorbidities), an important direction for ensuring fairness and clinical reliability.

Future work should conduct multi-site evaluations, extend to diverse data types, and integrate patient acuity scores. The CV-masking principle naturally extends to other clinical modalities: vital signs exhibit inherent volatility patterns, medication dosing shows patient-specific variation, and procedural timing reflects clinical urgency. Preliminary analysis suggests similar volatility-performance relationships exist across these domains, warranting systematic investigation.

\paragraph{Integration with Existing Models}
While our current implementation uses a bidirectional MAE architecture, the core principle of volatility-aware masking is architecture-agnostic and can be adapted to other EHR foundation model frameworks. CV-Masking is designed as a plug-in pretraining policy that operates on feature-level metadata, allowing it to modulate training emphasis regardless of the underlying model architecture.

Causal models like MOTOR \cite{steinberg2023motor} and ETHOS could incorporate CV-based principles through loss reweighting for high-volatility laboratory predictions during pretraining. Similarly, BEHRT's reconstruction losses for numeric laboratory values could be weighted by CV, and state-space models like EHRMamba could prioritize volatile measurements in their updates. The CV-based prioritization principle extends naturally across transformers, state-space models, and recurrent networks. Future work should explore extending CV-Masking to multi-modal EHR foundation models that jointly model diagnoses, medications, procedures, and laboratory tests, providing a comprehensive framework for clinically-informed pretraining across diverse architectures and data modalities.

\section{Conclusion}
\label{sec:conclusion}

We introduce CV-based masking, a volatility-aware pretraining strategy that systematically improves EHR foundation model performance by aligning masking probabilities with clinical reality. Our approach demonstrates that incorporating domain knowledge into self-supervised objectives can yield substantial improvements in both performance and efficiency.

The systematic improvements across 71\% of laboratory tests with large effect sizes (Cohen's d = 0.73), enhanced downstream clinical prediction performance, and 50\% reduction in training time establish CV-based masking as a principled pathway toward more robust and clinically meaningful EHR foundation models. Notably, the approach shows particular strength for clinically critical volatile markers including inflammatory proteins, liver enzymes, and hematologic indices. The perturbation analysis provides mechanistic evidence that this approach promotes deeper contextual learning, creating representations that better capture patient-specific temporal patterns.

As the field moves toward deployment-ready clinical AI systems, incorporating clinical knowledge into self-supervised objectives appears essential for developing models that are both technically sound and clinically meaningful. CV-based masking represents a concrete step toward this goal, demonstrating that thoughtful integration of medical domain expertise can substantially enhance foundation model capabilities.


\bibliography{references}

\clearpage
\appendix

\section{Training Efficiency Analysis}
\label{app:training}

Figure \ref{fig:training_curves} demonstrates the superior training efficiency of CV-based masking through actual training loss curves. CV-based masking converges in 33 epochs, representing a 50\% reduction compared to random masking (67 epochs) and 67\% reduction compared to variance-based masking (100 epochs). The CV-based training curve exhibits notably smoother convergence with reduced oscillations, indicating more stable gradient dynamics. This stability likely stems from the principled curriculum effect, where the model consistently focuses on appropriately challenging examples rather than oscillating between trivial and intractable predictions.

\begin{figure*}[htbp]
\floatconts{fig:training_curves}
{\caption{Training Convergence Analysis. CV-based masking achieves faster and more stable convergence than random and variance-based masking. These curves demonstrate that performance gains stem from the CV-masking strategy itself under controlled conditions, not from differential training durations or hyperparameter configurations.}}
{%
\begin{minipage}[t]{0.48\textwidth}
    \centering
    \includegraphics[width=0.95\textwidth]{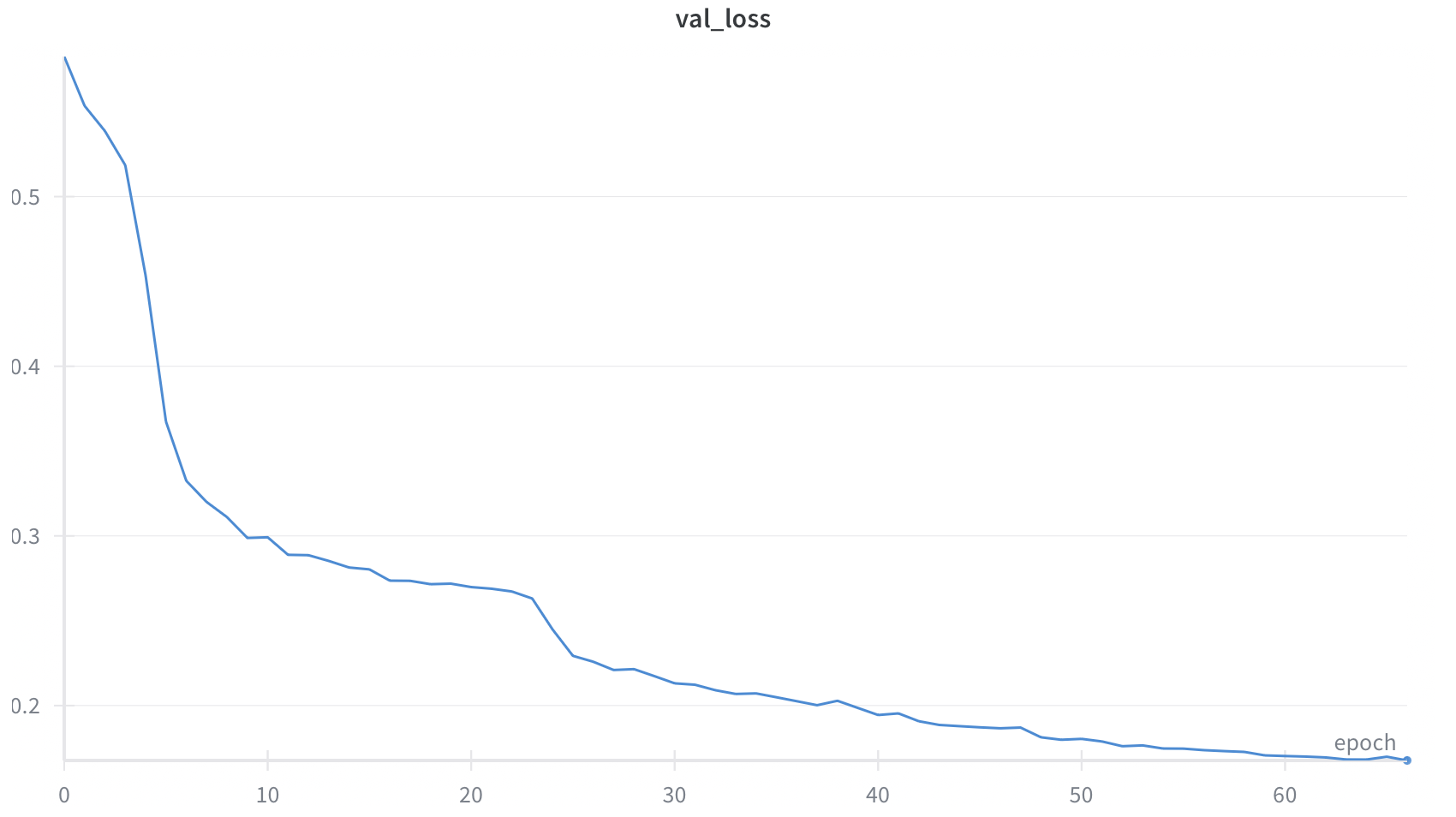}
    \vspace{1ex}
    Random Masking (67 epochs)
\end{minipage}%
\hfill
\begin{minipage}[t]{0.48\textwidth}
    \centering
    \includegraphics[width=0.95\textwidth]{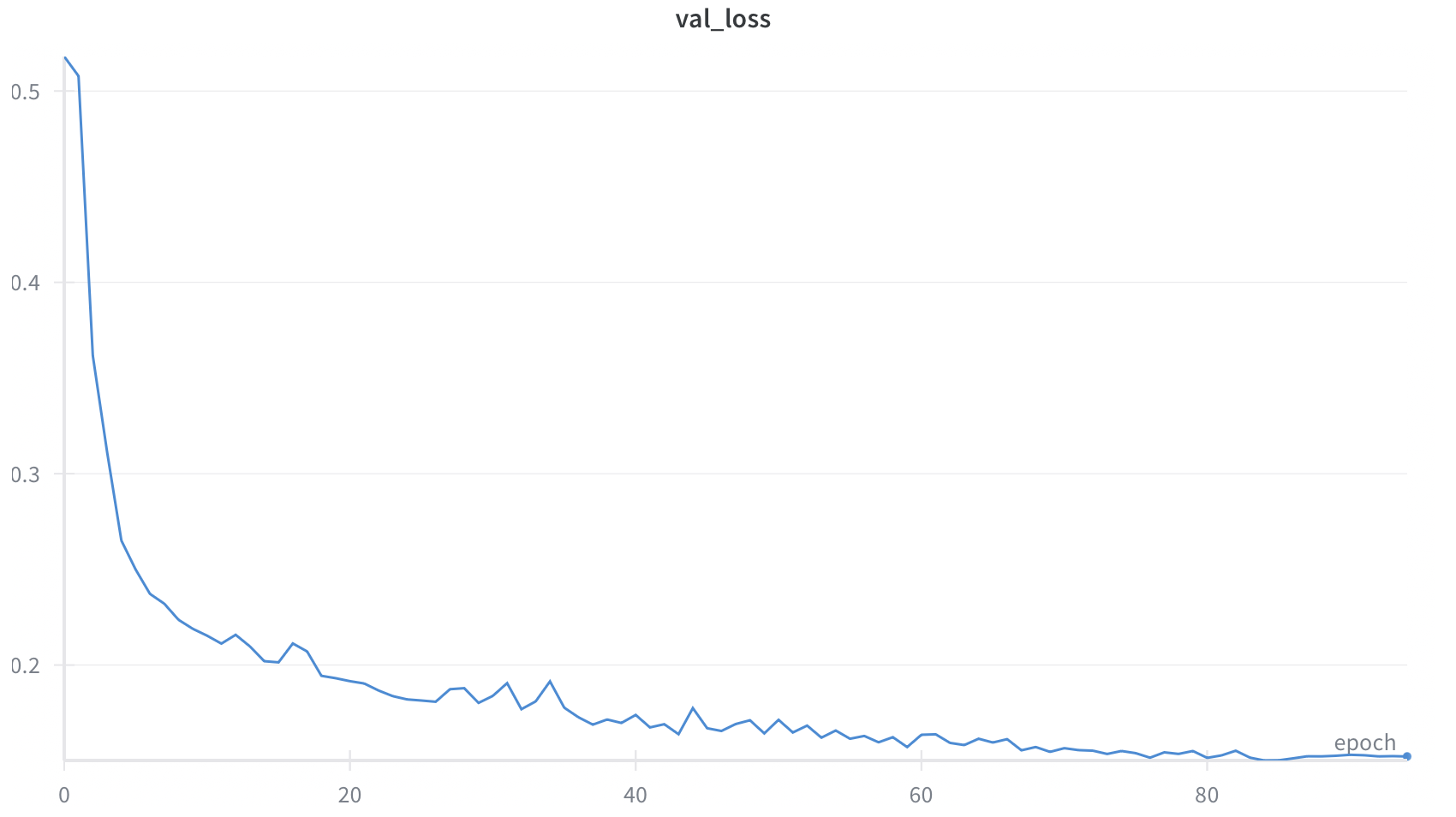}
    \vspace{1ex}
    Variance Masking (100 epochs)
\end{minipage}%

\vspace{1.2ex}

\begin{minipage}[t]{0.6\textwidth}
    \centering
    \includegraphics[width=0.95\textwidth]{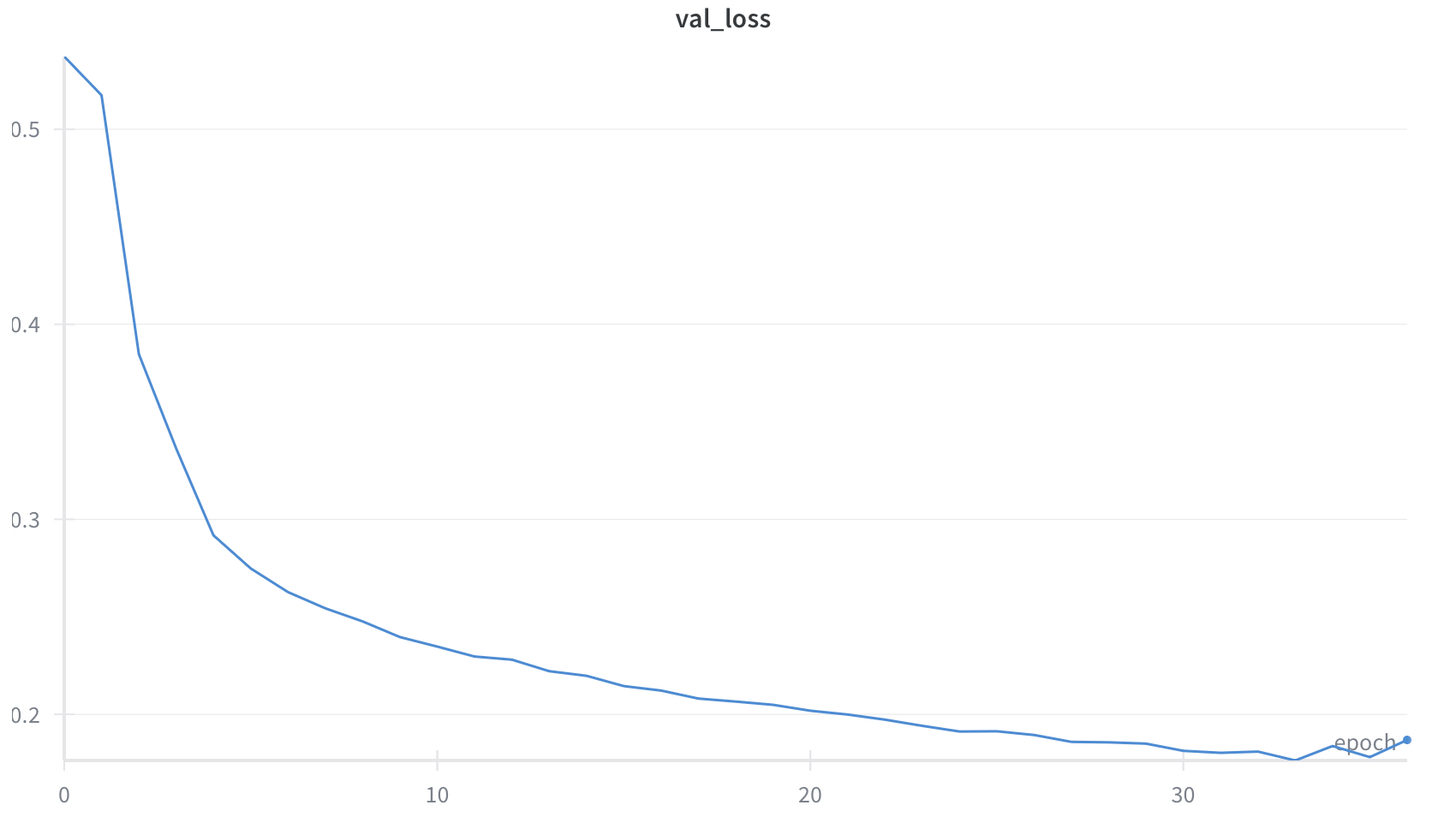}
    \vspace{1ex}
    CV-Based Masking (33 epochs)
\end{minipage}%
}

\end{figure*}

This efficiency gain stems from the principled curriculum learning effect where computational resources are allocated more effectively to challenging learning tasks, accelerating convergence while maintaining superior final performance. The contrast is particularly striking with variance-based masking, which shows erratic convergence patterns and requires 3× more epochs, highlighting the importance of scale-invariant volatility measures for effective curriculum design.

\section{Historical Context Learning Analysis}
\label{app:context}

To validate that CV-based masking promotes deeper contextual learning, we analyzed reconstruction performance as a function of available patient history. Figure \ref{fig:context_learning} demonstrates that CV-based masking consistently leverages historical information more effectively than baseline approaches, with steeper improvement slopes as more prior occurrences become available (left) and superior performance across different training frequencies (right). This provides mechanistic evidence that CV-based masking develops stronger reliance on contextual information rather than memorizing population-level statistics.

\begin{figure*}[htbp]
\floatconts{fig:context_learning}
{\caption{Historical Context Utilization. Left: Reconstruction error decreases as more prior laboratory occurrences become available, with CV-based masking showing steeper improvement slopes. Right: Performance across different training frequencies, demonstrating CV-masking's robustness for both common and rare laboratory tests.}}
{%
\begin{minipage}[t]{0.48\textwidth}
    \centering
    \includegraphics[width=\textwidth]{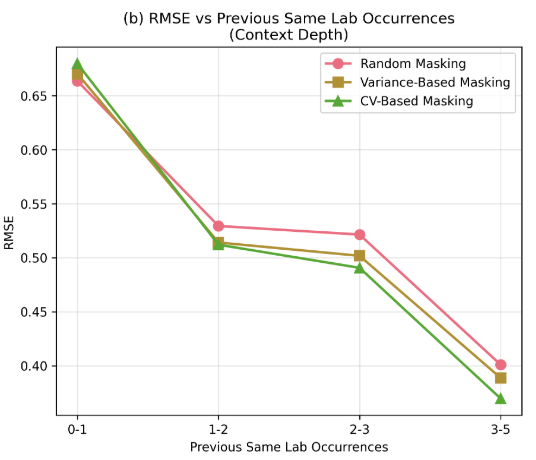}
    \vspace{1ex}
    Error vs. Prior Lab Occurrences
\end{minipage}%
\hfill
\begin{minipage}[t]{0.48\textwidth}
    \centering
    \includegraphics[width=\textwidth]{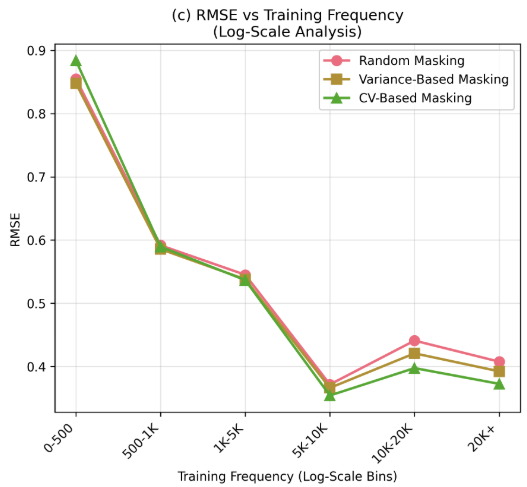}
    \vspace{1ex}
    Error vs. Training Frequency
\end{minipage}%
}

\end{figure*}

This contextual learning advantage directly supports our hypothesis that CV-based masking creates clinically meaningful representations by focusing on volatile biomarkers requiring sophisticated temporal modeling.

\section{Perturbation Methodology}
\label{app:perturbation}

To validate that CV-based masking learns meaningful temporal patterns, we corrupted historical laboratory values with adaptive Gaussian noise: $v'_i = v_i + \mathcal{N}(0, \sigma_{\text{adaptive}})$, where $\sigma_{\text{adaptive}} = \text{std}(v_{\text{prior}}) \times 0.6 \times 1.5$ ensures fair comparison across diverse laboratory types. Target predictions and temporal context were preserved. Performance degradation was quantified as relative MAE increase: $\frac{\text{MAE}_{\text{corrupted}} - \text{MAE}_{\text{original}}}{\text{MAE}_{\text{original}}} \times 100\%$.

CV-based models exhibited 2.1× greater degradation (9.8\% vs 4.7\%), providing causal evidence of stronger reliance on patient-specific temporal patterns across hematology, electrolyte, and metabolic markers.

\section{Per-Laboratory Reconstruction Examples}
\label{app:lab_examples}

To supplement the aggregate results in Section~\ref{sec:experiments}, this section provides visual examples of reconstruction performance on individual laboratory tests. The following figures show scatter plots of true versus predicted values under the three masking strategies.

\begin{figure*}[htbp]
    \centering
    \includegraphics[width=0.95\textwidth]{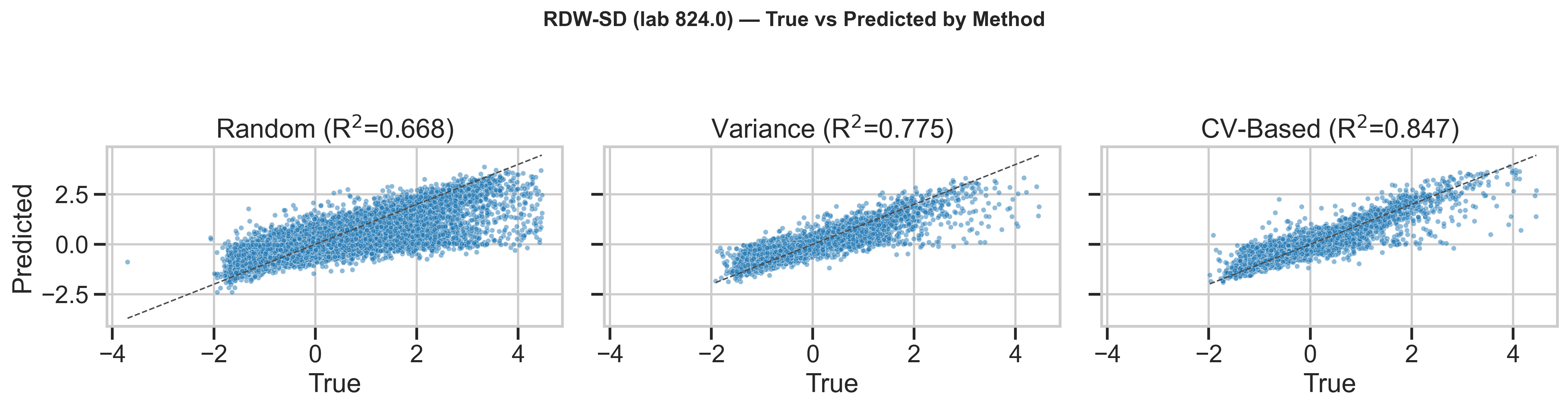}
    \caption{True vs. Predicted scatter plots for Red Cell Distribution Width (RDW-SD). Each point represents a single masked value prediction. The scatter becomes progressively tighter and more aligned with the ideal y=x line (dashed) as the masking strategy improves from Random (left) to Variance-based (center) to CV-Based (right). This visual improvement corresponds to R² increasing from 0.668 to 0.847.}
    \label{fig:rdw_r2}
\end{figure*}

\begin{figure*}[htbp]
    \centering
    \includegraphics[width=0.95\textwidth]{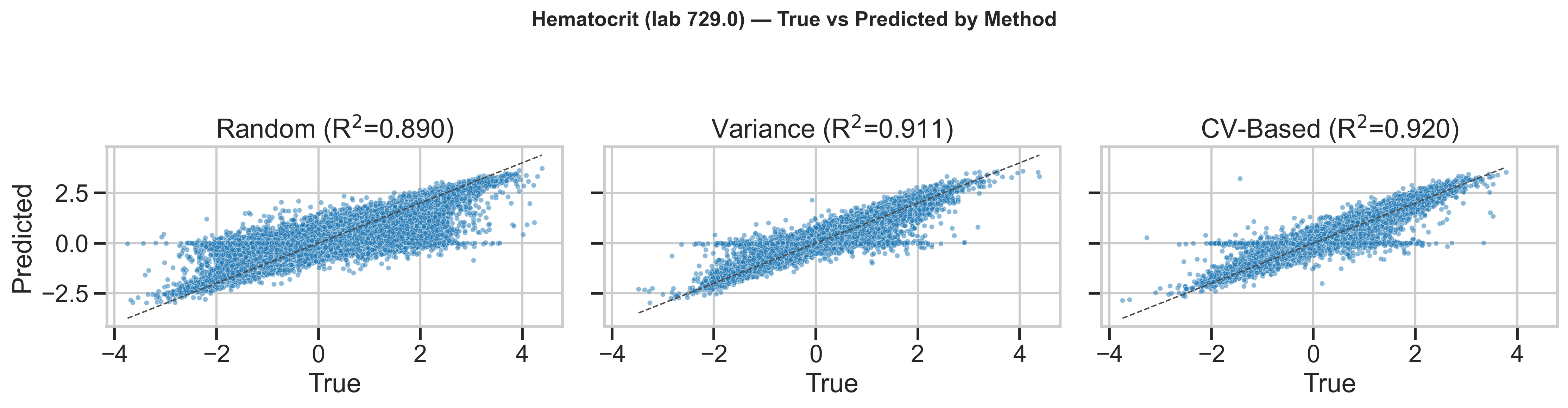}
    \caption{True vs. Predicted scatter plots for Hematocrit. As a stable, low-volatility biomarker, Hematocrit is predicted with high accuracy by all three methods. Progressive tightening of the scatter and R² increase from 0.890 (Random) to 0.920 (CV-Based) demonstrates the benefit of volatility-aware curriculum even on highly predictable labs.}
    \label{fig:hematocrit_scatter}
\end{figure*}

\begin{figure*}[htbp]
    \centering
    \includegraphics[width=0.95\textwidth]{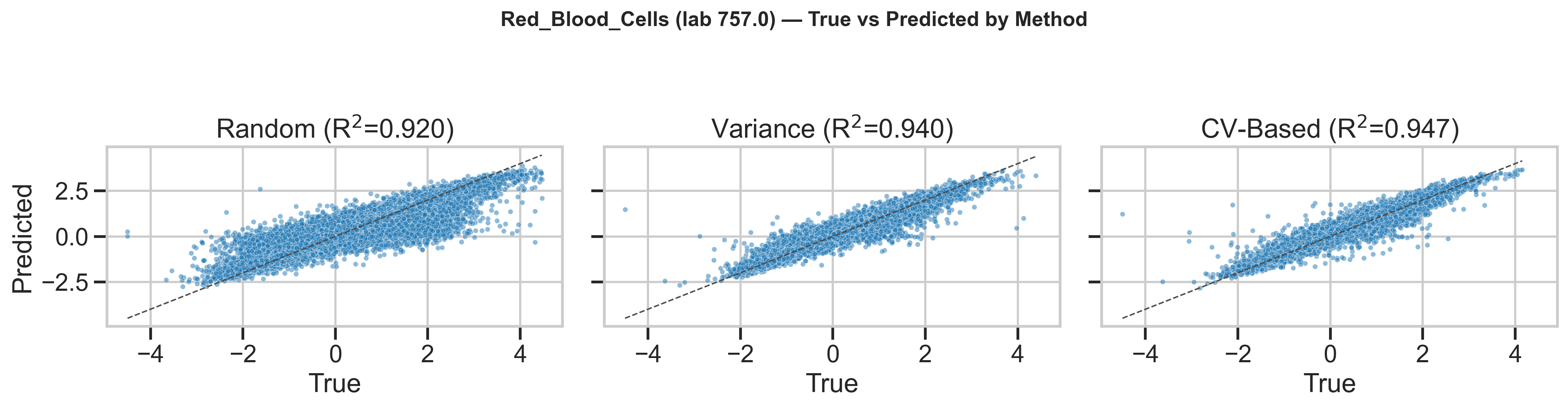}
    \caption{True vs. Predicted scatter plots for Red Blood Cells. Another example of a stable, low-volatility biomarker well-predicted by all methods. CV-Based approach (right) provides the best performance with R² of 0.947 and tightest clustering along the ideal y=x line.}
    \label{fig:rbc_scatter}
\end{figure*}

\FloatBarrier 

\section{Reproducibility Statement}
\label{app:reproducibility}

\subsection{Data}
\textbf{Dataset}: We use MIMIC-IV v3.1 \cite{johnson2023mimic}, a large critical care database. Data is structured using the MEDS (Medical Event Data Standard) format \cite{arnrich2024meds}, which encompasses diagnoses, procedures, medications, laboratory tests, and chart events. 

\textbf{Data Splits}: Patient-level 70/15/15 train/validation/test split ensuring no patient appears in multiple sets.

\subsection{Preprocessing}
Data preprocessing follows the MEDS (Medical Event Data Standard) pipeline \cite{arnrich2024meds, oufattole2024meds} available at: \url{https://github.com/mmcdermott/MEDS_transforms}

\textbf{Normalization}: Z-score normalization per laboratory test computed on training set

\textbf{Sequence Handling}: Maximum sequence length 512 events; longer sequences are truncated.

\subsection{Model Architecture}
\textbf{Encoder}: 8-layer Transformer
\begin{itemize}
    \item Hidden dimension (d\_model): 256
    \item Attention heads: 8
    \item Feed-forward dimension: 1024 (mlp\_ratio $\times$ embed\_dim = $4.0 \times 256$)
    \item Dropout: 0.1
    \item Additional: QK normalization enabled, layer scale 0.1, drop path rate 0.1
\end{itemize}

\textbf{Decoder}: 4-layer Transformer
\begin{itemize}
    \item Hidden dimension (d\_model): 128
    \item Attention heads: 4
    \item Feed-forward dimension: 512 (mlp\_ratio $\times$ decoder\_embed\_dim = $4.0 \times 128$)
    \item Dropout: 0.1
\end{itemize}

\textbf{Total Parameters}: Approximately 12-15M parameters (exact count depends on vocabulary size and embedding dimensions).

\subsection{Training}
\textbf{Optimization}:
\begin{itemize}
    \item Optimizer: AdamW ($\beta_1=0.9$, $\beta_2=0.95$, $\epsilon=10^{-8}$)
    \item Learning rate: $1 \times 10^{-4}$ with 1000-step warmup
    \item Weight decay: 0.01
    \item Batch size: 32 (held constant across all masking strategies for controlled comparison)
    \item Gradient clipping: max norm 1.0
    \item Mixed precision: FP16 (16-mixed)
\end{itemize}

\textbf{Masking}: 25\% masking ratio maintained across all strategies. For CV-Masking, codes with CV $>$ 75th percentile receive weight 0.8; codes with CV $\leq$ 75th percentile receive weight 0.2.

\textbf{Convergence Criteria}: Training terminates when validation loss does not improve for 10 consecutive epochs (early stopping with patience=10).

\textbf{Random Seeds}: 
\begin{itemize}
    \item Pretraining: Fixed seed 42 for reproducibility
    \item Downstream evaluation: Multiple runs with different seeds for robust evaluation
\end{itemize}

\subsection{Downstream Evaluation}
\textbf{Linear Probe Protocol}:
\begin{itemize}
    \item Encoder frozen (no gradient updates to pretrained weights)
    \item Single linear classification layer: 256 (encoder dim) $\rightarrow$ 1 (binary prediction)
    \item Training: AdamW optimizer, lr=$1 \times 10^{-4}$, 50 epochs, batch size 32
    \item Evaluation: AUROC and AUPRC metrics with confidence intervals computed via bootstrap (1000 samples)
\end{itemize}

\textbf{Task Definitions}:
\begin{itemize}
    \item In-ICU mortality: Binary prediction of death during ICU stay
    \item In-hospital mortality: Binary prediction of death during hospital admission
    \item 30-day readmission: Binary prediction of unplanned readmission within 30 days of discharge
\end{itemize}

\textbf{Training Time}: 
\begin{itemize}
    \item CV-based masking: Converges in 33 epochs
    \item Random masking: Converges in 67 epochs  
    \item Variance masking: Converges in 100 epochs
\end{itemize}

Full implementation details and code are available at \url{https://github.com/rajna-fani/meds-triplet-mae}.

\end{document}